\documentclass[10pt,letterpaper]{article}
\usepackage{cogsci}
\cogscifinalcopy 
\usepackage{pslatex}
\usepackage{apacite}
\usepackage{float} 
\usepackage{graphicx}

\title{Modeling Trust and Reliance with Wait Time in a Human-Robot Interaction}

\author{{\large \bf Akihiro Maehigashi (maehigashi.akihiro@shizuoka.ac.jp)} \\
  Shizuoka University, Shizuoka, Japan \\
\AND {\large \bf Seiji Yamada (seiji@nii.ac.jp)} \\
National Institute of Informatics, Tokyo, Japan\\ The Graduate University for Advanced Studies, SOKENDAI, Tokyo, Japan}

\begin{document}
\maketitle

\begin{abstract}
This study investigated how wait time influences trust in and reliance on a robot. Experiment 1 was conducted as an online experiment manipulating the wait time for the task partner's action from 1 to 20 seconds and the anthropomorphism of the partner. As a result, the anthropomorphism influenced trust in the partner and did not influence reliance on the partner. However, the wait time negatively influenced trust in and reliance on the partner. Moreover, a mediation effect of trust from the wait time on reliance on the partner was confirmed. Experiment 2 was conducted to confirm the effects of wait time on trust and reliance in a human-robot face-to-face situation. As a result, the same effects of wait time found in Experiment 1 were confirmed. This study revealed that wait time is a strong and controllable factor that influences trust in and reliance on a robot.

\end{abstract}

\section{Introduction}
Artificial intelligence (AI) has been pervasive in recent years, and it is considered to be inevitable in our lives. Social robots equipped with AI are also expected to be used to perform or assist human tasks. 

A successful human-robot interaction (HRI) has been considered to be developed by appropriately calibrating trust in a robot and adjusting one’s reliance (behavioral dependence) on the robot to maximize task performance \cite{Baker18, Lee04}. Trust is defined as ``the attitude that an agent will help achieve an individual's goals in a situation characterized by uncertainty and vulnerability'' \cite{Lee04}. 

The main factors affecting trust in a robot are robot-, human- and environment-related factors \cite{Hancock11}. In particular, the robot-related factor comprises performance-, appearance-, and behavior-related factors \cite{Khavas20}. Regarding the performance-related factor, there are studies which have shown how the reliability or competence of automated and robotic systems affect human trust \cite{Dzindolet02, Lee04, Wiegmann01}. Also, regarding the appearance-related factor, there are studies which have shown that greater levels of anthropomorphism increase trust in robots \cite{Christoforakos21, Kim20, Kopp22, Natarajan20}. Moreover, Regarding the behavior-related factor, there are studies which have shown that the robot's physical presence and behaviors increase trust \cite{Powers07}.

This study proposed that the wait time for a robot's actions as a performance-related factor can be a factor affecting trust in a robot and leading to proper use of it. People are known to be sensitive to wait times in interpersonal relationships \cite{Antonides02} and human-computer interaction (HCI) \cite{Nielsen93}. In particular, in HCI, a longer wait time for a computer response decreases user satisfaction extremely and increase stress \cite{Bouch00, Nah14, Nielsen93}. Therefore, there is a possibility that time spent waiting for a robot's actions influences trust and reliance.

\section{Related Work}

\subsection{Anthropomorphism in HRI}

Anthropomorphism has been known as a strong factor affecting trust. Anthropomorphism refers to the attribution of a human form, human characteristics, or human behavior to non-human entities \cite{Brian03}. Anthropomorphism is associated with the attribution of experience (emotional states such as fear, joy, pain, and consciousness) and agency (cognitive abilities such as thought, emotion recognition, communication, and planning.) \cite{Gray2007}.

\citeA{Natarajan20} showed that there was a positive correlation between the perceived anthropomorphism of the robot and trust in the robot. Also, \citeA{Kopp22} indicated that trust in the robot introduced with an emphasis on human-likeness was greater than that introduced with an emphasis on machine-likeness. Moreover, \citeA{Christoforakos21} revealed that robot's physical presence and behaviors increase perceived anthropomorphism and trust in a robot. 

Furthermore, Kim et al. \citeA{Kim20} showed that greater levels of anthropomorphism increase trust in a robot mediated by intelligence. In other words, a robot with greater levels of anthropomorphism is perceived as more intelligent.

\subsection{Wait time for Robot's actions}

Wait time for robot's actions can be considered a novel performance-related factor affecting trust in a robot. The performance-related factor, as one of the factors affecting human trust in HRI, is defined as a factor that determines “the quality of an operation performed by the robot,” such as its ability to achieve an operator’s goals or its faulty behavior \cite{Khavas20}. Wait time for robot's actions is related to the quality of an operation performed especially in time-critical tasks.

Also, in cognitive science, the task complexity has been measured by response time or task completion time as a representative value of cognitive cost. If a person takes a loner time duration for solving a task, the task is assumed to be more difficult increasing cognitive cost. Therefore, there is a possibility that a longer wait time for robot's action in a task could make people assume that the task is more difficult increasing its information processing cost. 

Many studies have shown that a longer wait time for a computer response in HCI decreases user satisfaction and increase stress \cite{Bouch00, Nah14, Nielsen93}. Most people are willing to wait for a computer response for only 2 seconds \cite{Nah14}. An acceptable wait time is considered to be 10 seconds, considered as the maximum response time before users lose their attention on a task \cite{Nielsen93}. Also, 15 seconds is considered tolerable, and most people give up on an on-going task with a wait time of more than 16 seconds \cite{Nielsen93}.

\section{Hypotheses}
\begin{figure}[tbp]
\centering
\includegraphics[width=\linewidth]{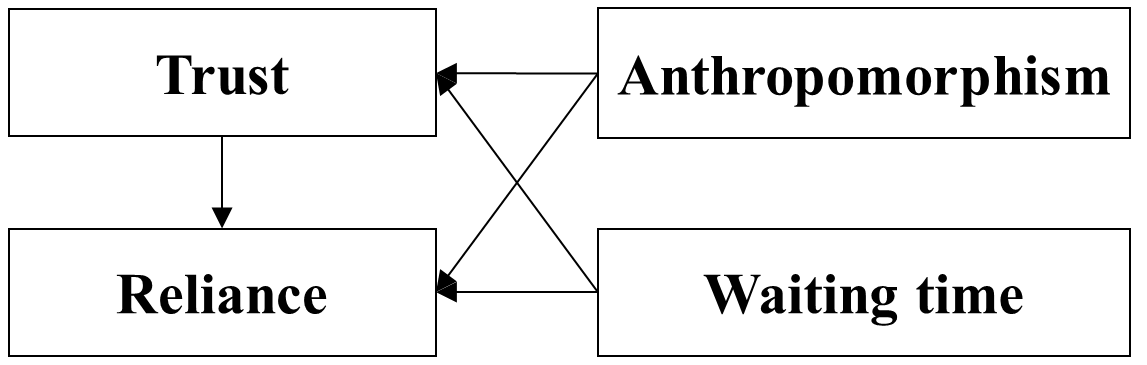}
\vspace{-4mm}
\caption{Hypothetical model of this study}\label{hypothetical-model}
\end{figure}

This study investigated how wait times for the partner's actions influence trust in and reliance on a task partner with consideration of the anthropomorphism of a partner. A hypothetical model of trust is shown in Figure~\ref{hypothetical-model}. 

First, in the model, trust is assumed to affect reliance. Trust models in human-automation interaction (HAI) and HRI have been developed showing that trust mediates and guides reliance \cite{Lee04} and also works as a reliance intention in which the user intend to rely on a robot \cite{Kim20}. Also, the trust-reliance relationship has been experimentally confirmed in many studies \cite{Dzindolet02, Lee1992trust, Wiegmann01} and has been applied and confirmed in HRI \cite{Christoforakos21, Kim20, Natarajan20}. 

Next, regarding anthropomorphism, on the basis of previous research showing that greater levels of anthropomorphism increase trust \cite{Christoforakos21, Kopp22, Kim20, Natarajan20}, positive effects from anthropomorphism on trust (\textbf{H1}) and reliance (\textbf{H2}) are expected to be confirmed. Also, regarding wait time, since a longer wait time for a computer response decreases user satisfaction and increases stress \cite{Bouch00, Nah14, Nielsen93}, the wait time could be considered to negatively influence trust (\textbf{H3}) and reliance (\textbf{H4}). 

Finally, since trust mediates and guides reliance \cite{Lee04}, the effect of anthropomorphism on reliance is assumed to be mediated by trust (\textbf{H5}). Also, the effect of wait time on reliance is also assumed to be mediated by trust (\textbf{H6}). 

We investigated the following hypotheses, especially focusing on the possibility of using the wait time to design trust.

\begin{description}

\item{\textbf{H1: The anthropomorphism of a partner will have a positive effect on trust.}}

\item{\textbf{H2: The anthropomorphism of a partner will have a positive effect on reliance.}}

\item{\textbf{H3: The wait time for a partner's action will have a negative effect on trust.}}

\item{\textbf{H4: The wait time for a partner's action will have a negative effect on reliance.}}

\item{\textbf{H5: The effect from the anthropomorphism of a partner on reliance is mediated by trust.}}

\item{\textbf{H6: The effect from the wait time for a partner's action on reliance is mediated by trust.}}

\end{description}

\section{Experiment 1}

\begin{figure}[b]
\centering
\includegraphics[width=\linewidth]{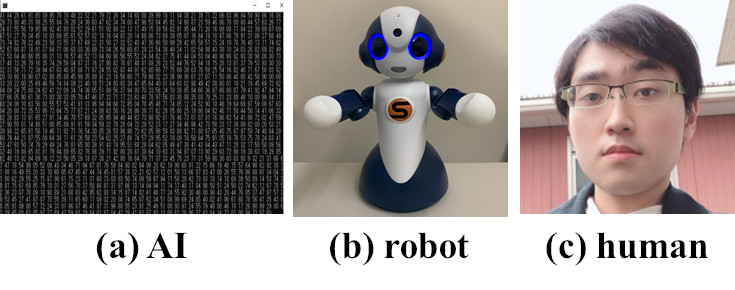}
\vspace{-4mm}
\caption{Task partner}\label{conditions}
\end{figure}

\begin{figure*}[tbp]
\begin{center}
\includegraphics[width=\textwidth]{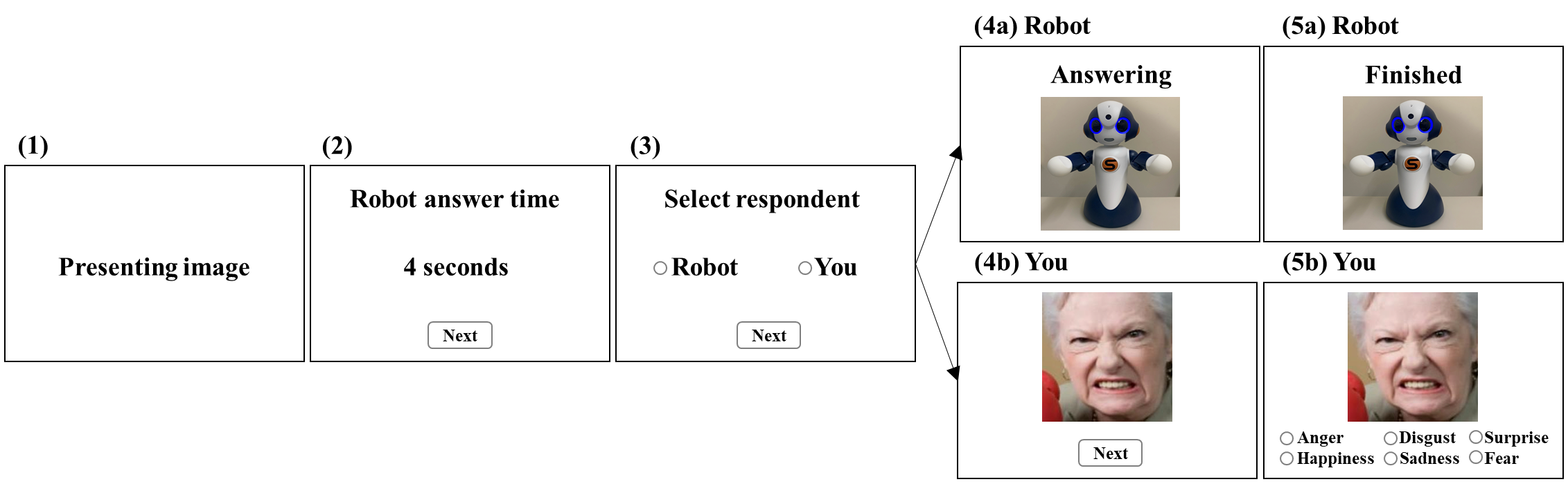}
\vspace{-4mm}
\caption{Task procedure in robot condition}\label{task-procedure}
\end{center}
\end{figure*}

\subsection{Experimental task}

The experimental task was set up to perform an emotion recognition task with a task partner having one of three types of anthropomorphic features, an AI, robot, or human, shown in Figure~\ref{conditions}. The task basically required participants to answer which of six emotions (anger, disgust, surprise, happiness, sadness, and fear) was expressed in pictures of human facial expressions using AffectNet \cite{Ali17}. The task procedure in the robot condition is shown in Figure~\ref{task-procedure}. 

\begin{enumerate}
\item[(1)] An image of a human facial expression was displayed to a task partner for 3 seconds.

\item[(2)] The partner indicated the estimated answer time (wait time), that is, how long it would take to answer the question. 
\item[(3)] On the basis of the estimated answer time, participants selected who would answer the question. 
\end{enumerate}

When the participants selected the robot partner as a respondent, (4a) the partner answered the question taking the same time duration as estimated and (5a) finished answering. A facial expression of a question and the answer of the partner were not displayed. \\

On the other hand, when the participants selected themselves, ``You'' as a respondent, (4b) an image of a question was displayed and (5b) the participants were required to answer the question, by clicking on one of the emotions.\\

In addition, while the partner answered and finished the question, one of the pictures in Figure~\ref{conditions} was displayed depending on the experimental condition. For the AI partner, an image of a command prompt displaying randomized numbers was used. For the robot partner, an image of Sota (Vstone Co., Ltd.) was used. However, in actuality, there were no partners answering the questions in this experiment, and the wait time was randomly chosen for each question. 

In this experiment, each participant answered a total of 40 questions with the partner. For the manipulation of the answer time (wait time), each number of seconds, from 1 to 20 seconds, was set to come up twice in randomized order. 
After each question was finished, participants were required to rate their trust levels in the partner. 

As in previous studies on human-automation interaction \cite{deVries03, Dzindolet03, Madhavan06}, participants were asked “how much do you trust your partner?” and were required to rate their trust levels in their partners on a 7-point scale (1: Extremely untrustable - 7: Extremely trustable). The trust level was measured before the start of each task and after each of four problems.

\subsection{Method}
\subsubsection{Experimental design and participants}
The experiment had a two-factor mixed-participants design. The factors were the anthropomorphism factor (AI, robot, and human) between participants and the wait time factor (from 1 to 20 seconds) within participants. 

An a priori statistical power analysis with G*Power \cite{Faul07} revealed that at least 68 data sets were needed for the multiple regression and the mediation analyses for testing the hypotheses with a medium effect size $(f = .15)$, power at .80, and alpha at .05. On the basis of this, we considered preparing data sets, explained later, and a total of 100 participants (81 male, 19 female) were recruited through a cloud-sourcing service provided by Yahoo! Japan. Their ages ranged from 19 to 75 years old ($M$ = 50.28; $SD$ = 11.80). They were randomly assigned to one of three conditions of the anthropomorphism factor. As a result, there were 33 participants in the AI, 34 in the robot, and 33 in the human conditions.

\subsubsection{Procedure.}
First, the participants gave informed consent, and after that, the emotion recognition task was explained. In the explanation, the task partner was introduced with one of the pictures in Figure~\ref{conditions} depending on the experimental condition. Participants were told that the partner was communicating in real time. Regarding the AI and robot partner, the AI or robot was explained as having an emotion recognition function. Also, regarding the human partner, the partner was introduced as one of the experimental collaborators waiting for the task. Participants were instructed to get as high a score as possible with the partner. They were also told that the partner did not always perform perfectly.

After that, participants performed a practice task with 5 questions with the answer time randomly selected from 1 to 5. After finishing the task, they rated the anthropomorphism of the partner using the Godspeed Questionnaire \cite{Bartneck09}. In the questionnaire, the participants rated the partner from 1 (fake) to 5 (natural) in Q1, from 1 (machine-like) to 5 (human-like) in Q2, from 1 (unconscious) to 5 (conscious) in Q3, from 1 (artificial) to 5 (lifelike) in Q4, and from 1 (moving rigidly) to 5 (moving elegantly) in Q5.

Also, in order to measure overall trust, participants answered the Multi-Dimensional Measure of Trust (MDMT), developed to measure trust in a robot \cite{Ullman18}. In this experiment, they were required to rate each of 4 words related to reliability (reliable, predictable, dependable, and consistent) and competence (competent, skilled, capable, and meticulous) on an 8-point scale (0: not at all - 7: very).

\subsection{Results}
First, there were participants who did not select a partner at all during the task. The data of these participants, 3 participants in the AI, 3 in the robot, and 5 in the human conditions, were eliminated from the analysis because they might have engaged in the experiment insincerely.

\begin{figure}[tbp]
\centering
\includegraphics[width=62mm]{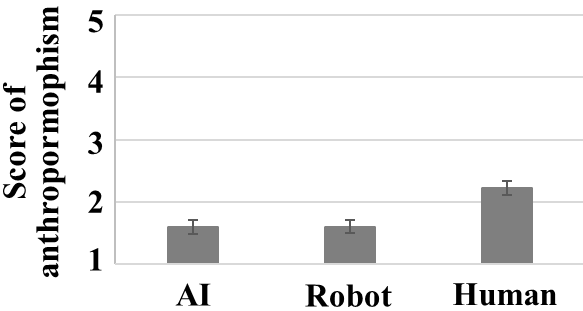}
\vspace{-4mm}
\caption{Scores of Anthropomorphism}\label{Anthropomorphism-score}
\end{figure}

Next, to confirm the manipulation of anthropomorphism, we conducted one-way ANOVAs on the average scores for anthropomorphism on the Godspeed Questionnaire (Figure~\ref{Anthropomorphism-score}). As a result, there was a significant main effect $(F(2, 86)=8.88, p < .01, \eta_{p}^2=.17)$. Multiple comparisons using Shaffer's modified sequentially rejective Bonferroni procedure showed that the score in the human condition was higher than those in the AI $(t(86)=3.68, p < .01, r=.37)$ and robot $(t(86)=3.67, p < .01, r=.37)$ conditions. However, there was no significant difference in the scores in the AI and robot conditions $(t(86)=0.04, p = ..97, r<.01)$. Even without the ratings for Q5 about movement, the same statistical differences were confirmed. Also, the average reliance rate, trust rating during the task, and MDMT rating in each condition were summarized in Table~\ref{result1}.

\begin{table}[tbp]
\begin{center} 
\caption{Average reliance rate, trust rating during the task, and MDMT rating of each condition in Experiment 1}\label{result1}
\vspace{-1 mm}
\small
\renewcommand{\arraystretch}{1.2}
\begin{tabular}{ cccc }
\hline
\textbf{Condition}  & \textbf{Reliance} & \textbf{Trust} & \textbf{MDMT}\\ \hline
\textbf{AI}  & 0.19 ($SD$=0.16)& 3.72 ($SD$=0.75)& 3.57 ($SD$=1.49) \\
\textbf{Robot}  & 0.28 ($SD$=0.20)& 3.30 ($SD$=0.75)& 3.87 ($SD$=1.06) \\
\textbf{Human}  & 0.22 ($SD$=0.23)& 3.66 ($SD$=0.78)& 3.31 ($SD$=1.40) \\
\hline
\end{tabular}
\end{center}
\end{table}

After that, we prepared data sets for testing the hypotheses. Each condition was divided into two groups depending on the order of the task performed. In each group of each condition, the average trust rating and reliance rate, the rate at which participants selected a particular partner, for each answer time were calculated. Also, regarding anthropomorphism, based on the results above, since there was a difference in anthropomorphism between the human condition and the other conditions, the AI and robot conditions were represented using ``0'', and the human condition was represented using ``1'' as dummy variables. Therefore, a total of 120 data sets was created. With these data sets, a satisfactory statistical power of .97 was achieved with a medium effect size $(f = .15)$ and alpha at .05

\begin{figure}[tbp]
\centering
\includegraphics[width=\linewidth]{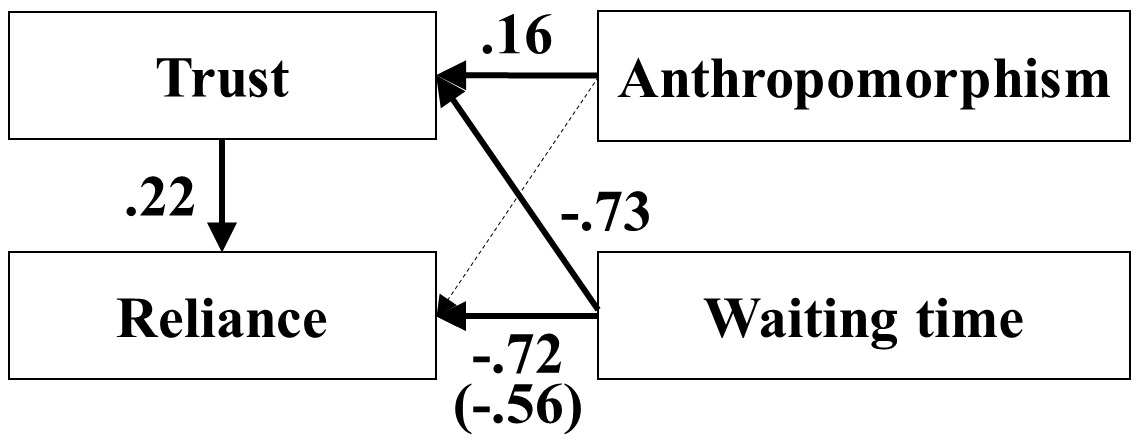}
\vspace{-4mm}
\caption{Summarized results for Experiment 1. Values are regression coefficients, which were statistically significant.}\label{result-exp1}
\end{figure}

The results are summarized in Figure~\ref{result-exp1}. First, we conducted a multiple regression analysis to investigate the effects of anthropomorphism and wait time on trust. The multiple regression model $({R}^2=.56)$ showed a significant positive effect of anthropomorphism on trust $(\beta=.16, p=.01)$ and a significant negative effect of wait time on trust $(\beta=-.73, p<.001)$. Second, we conducted the same analysis for the effects of anthropomorphism and wait time on reliance. The model $({R}^2=.53)$ showed a significant negative effect of wait time on reliance $(\beta=-.72, p<.001)$. However, there was no significant effect of anthropomorphism on reliance $(\beta=.12, p=.07)$. Finally, to test the mediation effect of trust from wait time on reliance, a mediation analysis with a Sobel test was performed. As a result, a significant mediation effect was revealed $(\beta=.22, t=2.39, p=.02)$. In addition, since there was no significant effect of anthropomorphism on reliance, a mediated analysis with anthropomorphism was not performed.

Furthermore, regarding the trust rating, we conducted correlational analyses between the average trust ratings during the task and MDMT ratings after the task. To perform the analysis, each rating was calculated for each participant. On the basis of the ratings, we performed correlation analyses between them in each condition. As a result, in the AI and human conditions, there were no significant correlations found between the trust and MDMT ratings $(AI: r = -0.11, p = .55, human: r = -0.16, p = .43)$. However, in the robot condition, there was a significant positive correlation between them $(r = 0.40, $p$ = .03)$.

\subsection{Discussion}
Regarding anthropomorphism, there was a positive effect of anthropomorphism on trust, and therefore, \textbf{H1} was supported. However, since there was no significant effect of anthropomorphism on reliance, \textbf{H2} and \textbf{5} were not supported. Moreover, regarding wait time, since there were negative effects of wait time on trust and reliance, \textbf{H3} and \textbf{4} were supported. Also, the effect of wait time on reliance was mediated by trust, and thus, \textbf{H6} was supported.

Regarding H1, this experiment manipulated a partner to answer the emotion recognition questions taking from 1 to 20 seconds. A longer answer time, that is, wait time, such as 20 seconds might have been unnatural especially for the human partner. Therefore, the participants might have given extremely lower ratings for the human partner, and the effect of anthropomorphism on trust might not have been found in this experiment. 

In addition, correlational relationships between trust ratings during the task and MDMT ratings after the task were only found in the robot condition. There is a possibility that some longer answer times might be impressive and unnatural for the human partner and intolerable for the AI partner, and consequently, some of the participants extremely decreased the MDMT ratings as overall trust.

\section{Experiment 2}
Experiment 2 was conducted to confirm the effects of wait time on trust and reliance, \textbf{H3}, \textbf{4}, and \textbf{6}, in a human-robot face-to-face situation. Since a robot has presence and physicality, the wait time might lead to different effects on trust.

\subsection{Method}
\subsubsection{Experimental design and participants.}
The experiment had a one-factor within-participants design. The factor was the wait time factor (from 1 to 20 seconds). A total of 19 Shizuoka University students (10 male, 9 female) participated in the experiment. Their ages ranged from 18 to 23 years old ($M$ = 20.06; $SD$ = 1.11).

\subsubsection{Experimental task and procedure.} 
\begin{figure}[tbp]
\centering
\includegraphics[width=80mm]{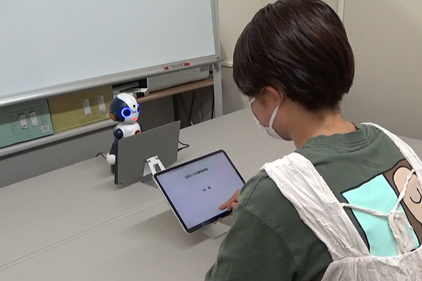}
\vspace{-1 mm}
\caption{Task situation in Experiment 2}\label{sota}
\end{figure}

The experimental task was basically identical to Experiment 1. However, in Experiment 2, a robot, Sota, was set in front of the participants with a tablet PC (Figure~\ref{sota}) and performed the task with the participants. Sota is a communication robot that enables natural dialogue using words with hand and body gestures. In Experiment 2, the following procedure was added to the procedure in Experiment 1.

\begin{enumerate}
\item[(1)] While an image of a human facial expression was displayed on the robot for 3 seconds, the color of the LED lights in the eyes changed from blue to yellow. After the 3 seconds, the color turned back to blue.  

\item[(2)] The estimated answer time (wait time) was displayed on the participant's tablet PC with the robot's auditory voice saying things such as ``the answer time is 4 seconds.'' While telling the time, the mouth blinked red along with the utterance.

\item[(3)] The procedure (3) was the identical to Experiment 1.

\item[(4)] When the robot was selected to answer, the color of the LED lights in the eyes changed from blue to green. The lights stayed green until the robot started to answer. Before answering a question, the lights turned back to blue. Also, while answering a question, the robot physically moved and tapped the tablet PC as if it were inputting an answer to the question. 

\end{enumerate}

In addition, there was no image of the robot displayed on the participant's tablet PC.

\subsection{Results}
First, there was one participant who did not wait for the robot's auditory voice indicating the answer time during the task. Also, there was one participant who kept tapping the tablet PC while the robot was answering the questions. In both cases, the robot could not behave accurately during the task. Therefore, the data of these two participants was eliminated from the analysis. A total of 340 data sets, 20 from each participant, was created. The average reliance rate and trust, MDMT, and anthropomorphism ratings were summarized in Table~\ref{result2}. After that, we prepared data sets. All the participants were divided into four groups in accordance with the order of tasks performed. For each group, data sets were created as in Experiment 1. Therefore, a total of 80 data sets was created. With these data sets, a satisfactory statistical power of .87 was achieved with a medium effect size ($f$ = .15) and alpha at .05.

\begin{figure}[tbp]
\centering
\includegraphics[width=\linewidth]{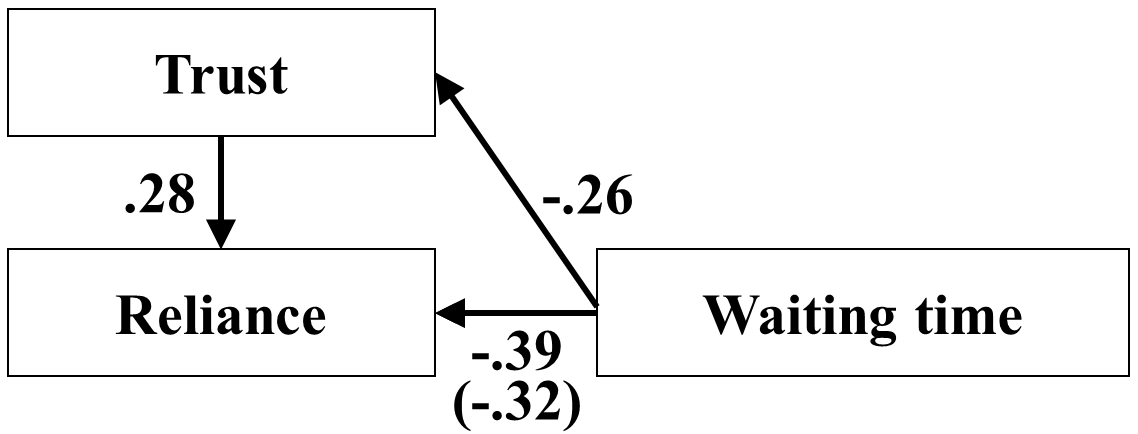}
\vspace{-4 mm}
\caption{Summarized results for Experiment 2. Values are regression coefficients, which were statistically significant.}\label{result-exp2}
\end{figure}

The results are summarized in Figure~\ref{result-exp2}. First, we conducted a simple regression analysis to investigate the effects of wait time on trust. The regression model $({R}^2=.07)$ showed a significant negative effect of wait time on trust $(\beta=-.26, p=.02)$. Second, we conducted the same analysis for the effects on reliance. The model  $({R}^2=.15)$ showed a significant negative effect of wait time on reliance $(\beta=-.39, p<.01)$. Finally, to test the mediation effect of trust from wait time on reliance, a mediation analysis with Sobel test was performed. As a result, a significant mediation effect was revealed $(\beta=.28, t=2.67, p=.01)$.

\begin{table}[tbp]
\begin{center} 
\caption{Average reliance rate and trust, MDMT, and anthropomorphism ratings in Experiment 2}\label{result2}
\vspace{+1 mm}
\small
\renewcommand{\arraystretch}{1.2}
\begin{tabular}{ cccc }
\hline
\textbf{Reliance} & \textbf{Trust} & \textbf{MDMT} & \textbf{Anthropomorphism}\\ \hline
0.34& 3.99& 4.71& 1.73 \\
($SD$=0.10)& ($SD$=0.74)& ($SD$=0.93) & ($SD$=0.65) \\
\hline
\end{tabular}
\end{center}
\end{table}

Furthermore, for the trust rating, we conducted correlational analyses between the average trust ratings during the task and MDMT ratings after the task. As a result, there was a significant positive correlation $(r = 0.64, p < .001)$.

\subsection{Discussion}
The experimental results confirmed \textbf{H3}, \textbf{4}, and \textbf{6} in a human-robot face-to-face situation. The same effects of wait time found in Experiment 1 were confirmed in Experiment 2. Therefore, the wait time was confirmed to be a strong factor influencing trust in and reliance on even a physically present robot. However, the R-squared values of the regression models were small. This might be because physical presence, body movements, or auditory sounds made the participants trust and rely on the robot differently, and thus, greater individual differences were considered to occur.

\section{General Discussion}
\subsection{Effects of wait time on trust and reliance}
This study investigated the influence of anthropomorphism and wait time on trust in and reliance on a task partner and developed a model of trust in a robot. As a result of Experiment 1 and 2, wait time was revealed to be a strong and controllable factor influencing trust in and reliance on a robot in HRI.

The results of Experiment 1 showed a possibility that some longer answer times might be unnatural for the human partner and intolerable for the AI partner as in HCI \cite{Bouch00, Nah14, Nielsen93}, and some of the participants extremely decreased the MDMT ratings as overall trust. However, such a phenomenon was not observed for the robot partner in Experiment 1 and 2. A robot is perceived to have less agency, such as thought, emotion recognition, communication, and planning, than human \cite{Gray2007}. Also, increasing anthropomorphism of AI and a robot induces human social behaviors \cite{Epley07}. Therefore, longer answer times, unnatural for a human partner and intolerable for an AI partner, might be valid and acceptable for a robot partner.

Also, although there was no statistical analysis conducted, the reliance, trust, and MDMT ratings in Experiment 2 were higher than those in Experiment 1. A physically present robot gives better impressions to people than a virtually displayed robot or animated character. In particular, people feel such robots to be more likable, helpful, enjoyable, trustworthy, and credible \cite{Powers07, Kiesler08}. Also, people become more compliant with a physically present robot than with a robot displayed on a screen \cite{Bainbridge11}. These effects were considered to occur because of the robot's physical presence in Experiment 2. In a such situation, the effects of wait time on trust and reliance were confirmed.

\subsection{Manipulation of wait time}

There are studies manipulating the wait time to form smooth and mutual coordination in HRI \cite{Maniadakis20, Someshwar13}. In these studies, a human and robot wait for each other's task completion to smoothly coordinate HRI. Contrarily, this study manipulated the wait time to influence trust in a robot and showed the possibility of using the time to design HRI. 

Trust could be lowered by simply increasing the wait time. Therefore, this could be done to suppress extreme over-trust. In particular, some people have a positive bias toward automated and robotic machines and tend to form over-trust \cite{Dzindolet02, Kim20}. For such users, the wait time could be set up to lower and calibrate trust in HRI. 

On the other hand, increasing trust by decreasing the wait time is technically difficult. However, since people feel the temporal duration to be longer when paying attention to time \cite{Zakey1997}, distracting attention away from time is a useful way to reduce the perceived wait time and increase trust. For HCI, progress bars \cite{Chris10, Asthana15, Kortum11} and auditory information \cite{Komatsu20} have been designed to catch users’ attention and reduce their perceived wait times.

\section{Conclusion}
This study investigated the influence of wait time and anthropomorphism on trust in and reliance on a task partner in a cooperative task and developed a model of trust for HRI. As a result of experiments, the anthropomorphism of a partner influenced only trust in the partner. However, the wait time for a partner's answer influenced trust in and reliance on the partner. Also, the effect of wait time on the reliance on the partner as mediated by trust was confirmed. The wait time was confirmed to be a strong and controllable factor influencing trust in and reliance on a task partner. There is a possibility that manipulating the wait time could contribute to trust calibration in HRI. 


\bibliographystyle{apacite}
\setlength{\bibleftmargin}{.125in}
\setlength{\bibindent}{-\bibleftmargin}

\bibliography{HRI23-1091}

\end{document}